\documentclass[sigconf, dvipsnames]{aamas}  % do not change this line!

\renewcommand\footnotetextcopyrightpermission[1]{} % removes footnote with conference information in first column
\usepackage{booktabs}

\usepackage{times}
\usepackage{helvet}
\usepackage{courier}
\usepackage{comment}
\usepackage{url}
\usepackage{bm}
\usepackage{amsmath}

\newtheorem*{remark}{Remark}

\usepackage{comment}
\usepackage{amssymb}
\usepackage{algorithm}
\usepackage{algpseudocode}
\usepackage{algorithmicx}
\usepackage{graphicx}
\usepackage{subfigure}
\usepackage{float}

\usepackage{flushend}

\begin{document}
\title{Robust Opponent Modeling via Adversarial Ensemble Reinforcement Learning in Asymmetric Imperfect-Information Games}  % put your title here!

\author{Macheng Shen}
%\authornote{Dr.~Trovato insisted his name be first.}
%\orcid{1234-5678-9012}
\affiliation{%
  \institution{Massachusetts Institute of Technology}
  \institution{Laboratory for Information \& Decision Systems}
  \streetaddress{77 Massachusetts Ave}
  \city{Cambridge} 
  \state{MA} 
  \postcode{02139}
}
\email{macshen@mit.edu}
\author{Jonathan P. How}
%\authornote{Dr.~Trovato insisted his name be first.}
%\orcid{1234-5678-9012}
\affiliation{%
  \institution{Massachusetts Institute of Technology}
  \institution{Laboratory for Information \& Decision Systems}
  \streetaddress{77 Massachusetts Ave}
  \city{Cambridge} 
  \state{MA} 
  \postcode{02139}
}
\email{jhow@mit.edu}

\begin{abstract}  % put your abstract here!
This paper presents an algorithmic framework for learning robust policies in asymmetric imperfect-information games (AIIG),  which are a subset of Bayesian games. We consider an information-asymmetric adversarial scenario in the context of multi-agent reinforcement learning (MARL), which has not been investigated before. In AIIG, the reward of a protagonist agent depends on the uncertain opponent type, which is a private information known only to the opponent itself. In contrast, the type of the protagonist agent is known to the opponent, which makes the decision-making problem difficult for the protagonist agent. For optimal decision-making, the protagonist agent has to infer the opponent type through agent modeling. We use multi-agent reinforcement learning  to learn opponent models through self-play, which captures the full strategy interaction and reasoning between agents. We apply ensemble training to make the learned protagonist policy robust against different opponents. We proposed a meta-optimization over the ensemble population, and demonstrated significantly improved robustness compared with baseline ensemble training without this meta-optimization, given the same computational budget. 

%However, agent policies learned from self-play can suffer from mutual overfitting. Ensemble training methods can be used to improve the robustness of agent policy against different opponents, but it also significantly increases the computational overhead. In order to achieve a good trade-off between the robustness of the learned policy and the computation complexity, we propose to train a separate opponent policy against the protagonist agent for evaluation purposes. The reward achieved by this opponent is a noisy measure of the robustness of the protagonist agent policy due to the intrinsic stochastic nature of a reinforcement learner. To handle this stochasticity, we apply a stochastic optimization scheme to dynamically update the opponent ensemble to optimize an objective function that strikes a balance between robustness and computation complexity. We empirically show that, under the same limited computational budget, the proposed method results in more robust policy learning than standard ensemble training.
\end{abstract}

  % put your semicolon-separated keywords here!

\maketitle

%%%%%%%%%%%%%%%%%%%%%%%%%%%%%%%%%%%%%%%%%%%%%%%%%%%%%%%%%%%%%%%%%%%%%%%%%%%%%%%%%%%%%%%%%%%%%%%%%%%%%%%%%
%% start of main body of paper

\section{Introduction}

\noindent Recent advances in deep reinforcement learning (DRL) have achieved breakthroughs in solving challenging decision-making problems in both single-agent environments \cite{dqn}, \cite{drqn}, \cite{inHandManipulation} and multiagent games \cite{alphaGo}, \cite{deepstack}, \cite{captureTheFlag}, \cite{openaiFive}, \cite{alphaStar}. Multiagent reinforcement learning (MARL) deals with multiple agents concurrently learning in a multiagent environment such as a multiagent game. One of the difficulties of learning in multiagent environments is that, in general, the state transition and reward depend on the joint action of all the agents. As a result, the best response of each agent depends on the joint policy of all the rest agents. This inter-dependency between agents makes it generally impossible to learn an optimal policy from a single agent prospect. In order to determine one's optimal policy, each agent has to reason about the likely policies of the other agents, and plans for its action accordingly, which is much more complicated than the single agent case.

Many of the successful MARL applications deal with two-player symmetric zero-sum games such as \cite{alphaGo}, \cite{deepstack}, \cite{neuralFictiousSelfPlay}, which was proved to have a Nash Equilibrium strategy profile that is equal to both the maximin strategy and the minimax strategy. This implies that the optimal policies in those games can be solved by using the worst case opponent policy. Since the game is symmetric, self-play is used which assigns one's own policy to its opponent, essentially reducing the multiagent learning problem to a single agent learning problem. 
 
In more general scenarios, there could be multiple equilibrium profiles. Agents do not necessarily adopt the equilibrium policy from the same equilibrium policy profile. As a result, solving for all the equilibrium profiles does not entail finding the optimal policies. Reasoning about other agents' policies becomes crucial for optimizing one's own policy. 

\subsection{Opponent modeling}
%, type-based reasoning, \cite{typeBased0}, \cite{typeBased1} 
Opponent modeling studies the problem of constructing models to reason about and make predictions about various properties (e.g. actions, goals, etc.) of the modelled agents. Classic methods, such as policy reconstruction \cite{policyReconst0}, \cite{policyReconst1} and plan recognition, \cite{planRecog0}, \cite{planRecog1} etc., develop parametric models to model agent behaviors \cite{agentModelingSurvey}. One of the limitations of these approaches is the requirement of domain-specific models, which could be difficult to acquire. Moreover, these models tend to decouple the interactions between the modeling agent and the modeled agents to simplify the modeling process, which is likely to be biased where strong coupling exists between agent rewards and interactions. In contrast, a more natural approach of opponent modeling is concurrently training all the agents via MARL in a self-play manner \cite{emergentOpenAISelfPlay}, \cite{maddpg}, \cite{modelingOthersUsingOneSelf}. This approach requires little domain-specific knowledge order than a black-box simulator. The interactions between the modeling agent and the modeled agents are fully captured in the joint observations, state-action pairs, and rewards. Moreover, concurrent learning provides a natural curriculum with the right level of difficulty for each agent \cite{emergentOpenAISelfPlay}.

\subsection{Ensemble training}

MARL in general-sum games, however, is more challenging than that in two-player zero-sum games. A general-sum game may have multiple equilibria, corresponding to a variety of diverse strong policies. During the training, the agents might have only seen a small subset of these  policies, which could lead to significant performance degradation when playing against unseen opponent policies. A common approach to mitigate this type of `policy over-fitting' is training policy ensembles such as in \cite{maddpg,pbt,captureTheFlag}. Each policy ensemble consists of several policies for each agent, which would be robust on average against all the policies within the ensembles of other agents. Ensemble training has also been widely applied to learning classifiers that are robust to adversarial attacks in computer vision \cite{ensembleTrainingCV}.

Although ensemble training improves the policy robustness, it also significantly increases the computational complexity, as each policy within an ensemble has to be optimized against ensembles of policies of other agents. In addition, choosing the right size of the ensemble is critical. A large ensemble size would likely result in high robustness but poor scalability, while a small size would scale better but potentially lead to a less robust policy. Therefore, finding a reasonable ensemble size that maintains a good trade-off between robustness and complexity is highly desirable, which has not been addressed in the related works. One fundamental issue with these works is the lack of a quantitative measure of robustness. Without this measure, we cannot optimize the ensemble size and the population selection. 

\subsection{Imperfect information and belief space planning}
%, \cite{factorPOMDP}
Imperfect-information and partial observability is another common difficulty in decision-making problems. Partially observable Markov decision process (POMDP) \cite{pomdp} decentralized-POMDP (Dec-POMDP) \cite{dec_pomdp} and Partially observable stochastic games (POSG) \cite{posg} are the decision-making models for single agent, multiagent fully cooperative and multiagent general sum scenarios, respectively. Model-based planning is the most prevalent technique for solving POMDP and Dec-POMDP (e.g., \cite{sarsop}, \cite{despot}, \cite{memoryboundedDP}). In POMDP, a value function satisfying Bellmen equation can be defined on the belief space. Piece-wise linear convexity (PWLC) is an important property of finite horizon POMDP value functions \cite{pomdp}. This property implies that belief state of high uncertainty has lower value while that of low uncertainty has higher value. Emergent exploration behavior is a natural result of PWLC.

\subsection{Recurrent policy}

Belief space policy, where the belief is a sufficient statistics of the action-observation history, is a special case of recurrent policy. One limitation of belief space planning is the requirement of an environment model for belief update, which is typically unavailable or intractable in scenarios with complex environments. In model-free DRL, recurrent neural network (RNN) is a widely used architecture to handle partial observability. Although RNN-based DRL approaches have achieved impressive successes in partially observable domains (e.g., \cite{captureTheFlag}, \cite{openaiFive}, \cite{alphaStar}), we identified two limitations of model-free learning with RNN: First, learning exploration behavior could be challenging. To the best of the the authors' knowledge, there is little evidence in literature showing emergent exploration behavior learned by RNN alone. Our conjecture is that RNN has to simultaneously learn an encoding of the action$-$observation history that has a similar information structure as the belief space, and a mapping from this hidden encoding to an optimal action. This is a more challenging learning task that a single black-box RNN might struggle to accomplish, as compared with model-based planning. Second, since there is no belief state in the RNN approach, RNN policy learns directly from the actual reward instead of the belief space reward. The actual reward could be very noise due to different realizations of the hidden state. This high reward variance poses challenges to reinforcement learning algorithms.

Besides, within the imperfect information and partially observable domains, different problems have different levels of difficulty. Most works deal with domains where the hidden state has a well-modeled probabilistic relationship with the observations, such as partial observability due to sensor noise or failure \cite{pomcp_sensorfailure}, limited field of view \cite{limitedfov}, screen flickering \cite{drqn}. These types of partial observability are relatively simple, in the sense that the hidden information can be inferred without bias via Bayes' rule. In the rest cases, the hidden state cannot be directly inferred from the observation. For example, in Poker game, the observation is all the hands that have been played, and the hidden state is the hands that have not been revealed. There is no probabilistic relationship between the hidden state and the observation. Nonetheless, inferring the hidden state is still possible given knowledge about the agent types and assumptions about rationality (agent modeling). For example, in Bayesian game theory, a Bayesian-Nash Equilibrium is well-defined given a joint equilibrium policy profile assuming perfect rationality \cite{learningBayesianGame}. However, this belief is likely to be biased since it is unlikely that the actual agent adopts the exact model policy. Moreover, in even more complicated imperfect-information scenarios, agent types could also be uncertain. For example, in one-night werewolf game, \cite{werewolf} agents do not know whether the other agents are their ally or enemy. In this case, one has to jointly reason about the (hidden) agent types and their policies, which is more challenging than the aforementioned situations.

\section{Overview and our contributions}

This paper presents an algorithmic framework for learning robust policies in asymmetric imperfect-information competitive games. We mainly focus on the scenarios where the opponent type is unknown to the protagonist agent but the joint reward is strongly correlated with this hidden type. This setting models a spectrum of real world scenarios, but has seldom been studied in the context of MARL. 

We use self-play with policy ensembles to learn a population of opponent models. We adapt the cooperative-evolutionary reinforcement learning (CERL) approach \cite{cerl} from single agent reinforcement learning to multi-agent settings for learning diverse opponent models. Diversity within the opponent ensemble is crucial for robust learning of protagonist agent policy. We apply policy distillation to synthesize the learned opponent policy ensemble for explicit belief update via Bayesian rule. We empirically show that learning an explicit belief space policy outperforms RNN-based approach. In order to obtain a good trade-off between policy robustness and complexity due to ensemble training, we propose to train a separate evaluation policy, which is optimized against the learned protagonist agent policy. The value of the evaluation policy is interpreted as a noisy measure of robustness, which constitutes the objective function of a stochastic discrete optimization over the power set of the policy ensemble. We apply simulated annealing to dynamically optimize the opponent policy set for an optimal trade-off between robustness and complexity. The resulting protagonist policy is empirically shown to be significantly more robust than that learned without this meta-optimization step, given same computation budget. 

The key contributions of this work are summarized as follow:

\begin{enumerate}
	\item We propose the first (to the best of our knowledge) MARL approach with belief state for solving a subset of Bayesian games, which we refer to later on as Asymmetric Imperfect-Information Games (AIIG). This is analogous to reinforcement learning within belief space applied to single agent POMDPs.  
	\item We identify one of the key challenges of solving AIIG, opponent modeling, which is analogous to the environment modeling in single-agent POMDPs. We derive a general formula for inferring opponent's hidden type, which reduces to learning an opponent model. We demonstrate the necessity of opponent modeling and belief space reasoning by showing that it significantly outperforms RNN-based approaches without explicit opponent modeling.
	\item We elaborate on this opponent modeling paradigm by adopting one of the state of the art ensemble training approach (CERL) for robustness, while also increases the complexity significantly. We thereby propose a meta-optimization scheme that improves the effectiveness of ensemble training for reducing complexity.
\end{enumerate}

\section{Preliminary}
In this section, we review the preliminary of the decision-making framework and solution techniques.

\begin{comment}
\subsection{Partially observable stochastic games}

A partially observable stochastic game (POSG) is a tuple $\langle\mathcal{I}, \mathcal{S}, \{b^0\}, \{\mathcal{A}_i\}, \{\mathcal{O}_i\}, \mathcal{P}, \{R_i\}\rangle$, where,
\begin{itemize} \itemsep -0.025in
    \item $\mathcal{I}$ is a finite set of agents indexed by $1, \ldots, n$
    \item $\mathcal{S}$ is a state space
    \item $b^0 \in \Delta(\mathcal{S})$ is the prior state distribution
    \item $\mathcal{A}_i$ is the action space of each agent, and we use $\boldsymbol{a}=\left\langle a_{1}, \dots, a_{n}\right\rangle$ to denote the joint action
    \item $\mathcal{O}_i$ is the observation space for each agent, and we use $\boldsymbol{o}=\left\langle o_{1}, \dots, o_{n}\right\rangle$ to denote the joint observation
    \item $\mathcal{P}$ is the Markovian state transition and observation probability, which is denoted as $\mathcal{P}(s'|s, \boldsymbol{a})$ and $ \mathcal{P}(\boldsymbol{o}|s, \boldsymbol{a})$
    \item $R_{i} : \mathcal{S} \times \overrightarrow{\mathcal{A}} \rightarrow \Re$ is the reward function of each agent
\end{itemize}
\end{comment}

\subsection{Bayesian Games}

A Bayesian Game (BG) is given by $ G = \langle\mathcal{I}, \langle \mathcal{S}, \mathcal{H} \rangle, \{b^0\}, \{\mathcal{A}_i\}, \{\mathcal{O}_i\},$ $\mathcal{P}, \{R_i\}\rangle$, where,
\begin{itemize} \itemsep -0.025in
    \item $\mathcal{I}$ is a finite set of agents indexed by $1, \ldots, n$
    \item $\Omega = \langle \mathcal{S}, \mathcal{H} \rangle$ is the set of state of nature, which includes the physical states and the agent hidden states corresponding to agent types in our problem
    \item $b^0 \in \Delta(\mathcal{S \times \mathcal{H}})$ is the common prior probability distribution over $\Omega$
    \item $\mathcal{A}_i$ is the action space of each agent, and we use $\boldsymbol{a}=\left\langle a_{1}, \dots, a_{n}\right\rangle$ to denote the joint action
    \item $\mathcal{O}_i$ is the observation space for each agent, and we use $\boldsymbol{o}=\left\langle o_{1}, \dots, o_{n}\right\rangle$ to denote the joint observation
    \item $\mathcal{P}$ is the Markovian state transition and observation probability, which is denoted as $\mathcal{P}^T(s'|s, \boldsymbol{a})$ and $ \mathcal{P}^O(\boldsymbol{o}|s, \boldsymbol{a})$
    \item $R_{i} : \Omega \times \overrightarrow{\mathcal{A}} \rightarrow \Re$ is the reward function of each agent
\end{itemize}

\subsection{Bayesian-Nash Equilibrium}
A Bayesian-Nash Equilibrium (BNE) is a joint strategy profile such that none of the agents could increase its expected reward (with respect to its own belief) by unilaterally deviating from such joint strategy, where the belief update is based on this strategy profile. This means that the belief-based strategy and the belief update rule are closely coupled, which makes it much more difficult to solve than games with perfect information. In a reinforcement learning context, this BNE solution concept suggests solving for a belief-space policy with the belief update rule induced from the opponent's (approximately) optimal policy (opponent modeling).  

\subsection{Asymmetric imperfect-information game}
%\XX{why? is that an important sub game to look at?}
In this paper, we primarily focus on a special subset of BG, which we refer to as asymmetric imperfect-information game (AIIG). We define AIIG as BG where there is no uncertainty over the type of the protagonist agent, while the opponent's type is hidden to the protagonist agent. This information-asymmetry adds significant difficulty to the protagonist agent's decision-making, and it has to reason about the opponent's type from its own observation.

AIIG models some important real world scenarios. In a buyer-seller game, the seller knows the true value of the goods while the buyer does not, which leads to different initial belief over the value of the goods. In an urban-security scenario, suppose a police officer wants to identify a terrorist among a swarm of people. The officer does not have prior knowledge over the type of each person, so he has to assign the same belief to each of the people. In contrast, the terrorist knows the type of all the other innocent, conditioned on its own type.

\begin{figure*}[t]
  \centering
  \includegraphics[trim=0 0 0 2,clip, width=\textwidth]{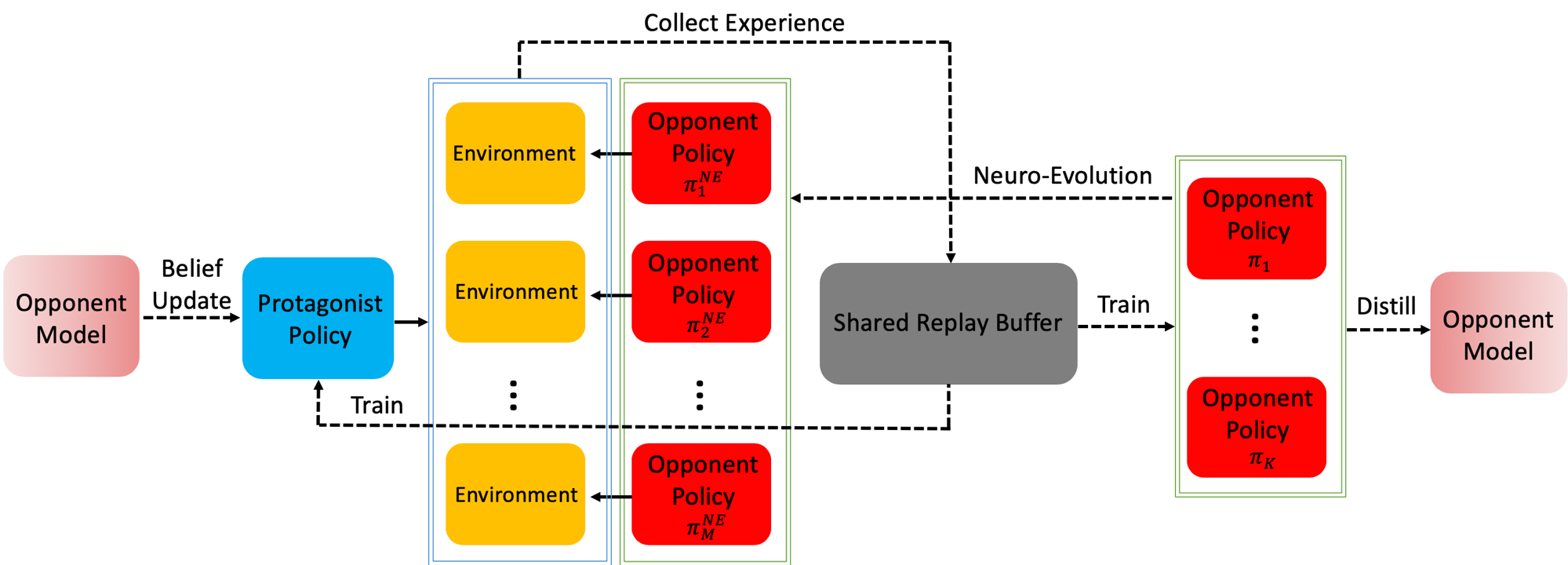}
  \caption{Illustration of the workflow: we train one protagonist policy that uses an internal opponent model for belief update. The opponent model is learned by distilling an ensemble of opponent policies trained against the protagonist policy. Both the protagonist and the opponent improve their skills through this self-play}
  \label{illustration}
\end{figure*}

\begin{comment}
\subsection{POSG with extended observation space}

We propose a small modification to the POSG framework, which we refer to as POSG with extended observation space (POSGEOS). In POSG, one of the general assumptions is that agent cannot observe the observations and actions of the other agents. We propose to extend the observation space with the actions taken by the other agents. We argue that this is a realistic assumption in many real game scenarios. For example, in the buyer-seller game, the actions are the prices proposed by the buyer or the seller, which is observed by both agents. In games with deterministic state transition, often the actions taken by agents could be deduced from the state transition. Effectively, the actions are also observable in these games. 

The motivation of proposing POSGEOS is for tackling imperfect-information scenarios with uncertain agent types. The idea is to infer the hidden types of the other agents from observing their actions, by reasoning about their policies based on some rationality assumption. 

\end{comment}

\subsection{Belief space reward}
In single agent partially observable domains, value function is defined as the expected cumulative reward with respect to the state-action distribution under the belief space policy $\pi$,
\begin{equation}
\begin{aligned}
    V^{\pi}(b_{0}) = & \sum_{t=0}^{\infty} \gamma^{t} \sum_{s^t} p(s^t) r(s^{t}, a^{t})\\
    = & \sum_{t=0}^{\infty} \gamma^{t} \mathbb{E}_{s^t\sim p(s^t), a^t \sim \pi(b^t)}[R(s^{t}, a^{t})],
\end{aligned}
\label{value_func}
\end{equation}
where $p(s^t)$ is the state distribution, and $b^t$ is the belief over state. If the belief is unbiased, then $p(s^t) = b(s^t)$, and Eq. \ref{value_func} degenerate to 
\begin{equation}
    V^{\pi}(b^{0})=\sum_{t=0}^{\infty} \gamma^{t} r(b^{t}, a^{t}), 
    %\text{where} \mbox{ }
    %r(b, a)=\sum_{s \in S} b(s) R(s, a),
\label{belief_space_reward}
\end{equation}
where $r(b,a) = \mathbb{E}_{s\sim b(s)}[r(s,a)]$ is the belief space reward. In reinforcement learning, we sample reward from the environment. The belief space reward sample $r(b^t, a^t)$ clearly has lower variance than the actual reward sample $r(s^t, a^t)$, because the uncertainty associated with the state distribution has been analytically marginalized out. As a result, learning in the belief space benefits from the low reward variance, in contrast to RNN-based approaches that learn directly from state space reward which has higher variance. 

In general, however, the state distribution $p^t$ and the belief $b^t$ could be different, for example, when the environment model $\mathcal{P}$ used for belief update is biased. In this case, the policy maximizing the belief space cumulative reward Eq. \ref{belief_space_reward} does not necessarily maximize the actual cumulative reward Eq. \ref{value_func}. That is, the agent learns an optimal policy in its imagined world, which  is actually sub-optimal due to the discrepancy between its world model and the actual world. This makes it challenging to solve asymmetric imperfect-information games with uncertain opponent types. On one hand, we want to exploit belief space reward for stable learning. On the other hand, however, belief update requires an opponent model, which is likely to be biased. Therefore, accurately modeling the opponent is crucial in our problem.

\section{Approach}
We first give an overview of our approach. We use MARL for policy learning, where competitive agents are trained against each other to consistently improve their skills. We use neural network to represent a belief space policy that maps a belief over the hidden state to an action. The belief state is updated via Bayes' rule using a learned model of the opponent policy. The opponent model learning process consists of an ensemble policy training step and a policy distillation step. We apply a neuro-evolutionary method to improve the diversity of the ensemble population for robustness. The above steps are illustrated in ig. \ref{illustration}. We then developed a stochastic optimization framework to meta-optimize the policy ensemble allocation for improved balance between robustness and complexity. We present the detail of each step in the following sections.

\subsection{MARL with ensemble training}
In order to improve the policy robustness of the protagonist agent, we formulate its RL objective as the average cumulative reward against an ensemble of opponent policies of size $K$, as in \cite{maddpg},
\begin{equation}
    J\left(\pi_{i}\right)=\mathbb{E}_{k \sim \operatorname{unif}(1, K), a_i \sim \pi_{i},  a_{-i} \sim \pi_{-i}^{(k)}}\left[\sum_{t=0}^{\infty} \gamma^t r_i(s, \boldsymbol{a})\right],
\label{RL_objective}
\end{equation}
where the policy ensemble $\{\pi_{-i}^{(k)}, k = 1, 2,..., K\}$ is also learned from training RL agent against the protagonist policy. Via this self-play, both the protagonist agent and its opponent improve their policies. Nonetheless, there is no explicit mechanism to enforce distinction among the policies within the ensemble. As a result, there could be redundant policies that are very similar to the others. 

To address this redundancy issue, we apply the cooperative evolutionary reinforcement learning (CERL) approach \cite{cerl}. The key idea is to use different hyper-parameter settings for each opponent policy, while use an off-policy learning algorithm and a shared experience replay buffer to keep the advantage of concurrently training multiple policies. Furthermore, neuro-evolutionary algorithm is applied to create mutated policies from the ensemble, and the trajectory under the mutated policies are also stored in the share experience replay buffer for better diversity and exploration.

\subsection{Belief space policy and belief update}

In the asymmetric imperfect-information games, however, the global state is not fully observable. We use the belief space approach for agent policy learning. Agents explicitly maintain a belief over the hidden states (e.g. hidden state includes the actual opponent types), and learns a belief space policy that maps belief to action. We parameterize this mapping using a multi-layer perceptron (MLP). The learning objective, instead of Eq. \ref{RL_objective}, now becomes,
\begin{equation}
    J\left(\pi_{i}\right)=\mathbb{E}_{k \sim \operatorname{unif}(1, K), a_i \sim \pi_{i}(b_i),  a_{-i} \sim \pi_{-i}^{(k)}}\left[\sum_{t=0}^{\infty} \gamma^t r_i(b_i, \boldsymbol{a})\right].
\label{RL_objective}
\end{equation}
A belief update mechanism is required to fully specify the agent policy. The belief is the posterior distribution over the hidden states given action and observation history, $b^t_i = p(s^t, h^t|o_i^{0:t})$.

Using Bayesian rule, we can write down the following equation,

\begin{equation}
	b_i^t \propto p(o^t_{i}| s^t, h^t, o^{0:t-1}_i) p(s^t, h^t|o_i^{0:t-1})
\label{belief_update}
\end{equation}

We further simply Eq. (\ref{belief_update}), the first term is

\begin{equation}
\begin{aligned}
	p(o^t_{i}| s^t, h^t, o^{0:t-1}_i) & = p(o^t_{i}| s^t, h^t) \\
	& = \int p(o^t_{i}| \boldsymbol{a}^t, s^t, h^t) p(\boldsymbol{a}^t|s^t, h^t) d\boldsymbol{a}^t
\end{aligned}
\label{first_term}
\end{equation}

where the first term in Eq. (\ref{first_term})$p(o^t_{i}| \boldsymbol{a}^t, s^t, h^t)$ is the observation probability. It is reasonable to assume $p(o^t_{i}| \boldsymbol{a}^t, s^t, h^t) = p(o^t_{i}| \boldsymbol{a}^t, s^t) = \mathcal{P}^O(o^t_{i}| \boldsymbol{a}^t, s^t)$, whose interpretation is that agents' observations are only dependent on physical states and actions, and are not affected by their internal type states. 

The second term in Eq. (\ref{first_term}) $p(\boldsymbol{a}^t|s^t, h^t)$ is the key connection between opponent type inference and opponent policy modeling. Intuitively, this term is closely related to agent policy, as can be seen by introducing the joint observation immediately before all the agents taking actions, ${o}^{t^{-}}$.  $p(\boldsymbol{a}^t|s^t, h^t) = \int p(\boldsymbol{a}^t|\boldsymbol{o}^{t^{-}}, s^t, h^t) p(\boldsymbol{o}^{t^{-}}|s^t, h^t) d\boldsymbol{o}^{t^{-}}$. The second term $p(\boldsymbol{o}^{t^{-}}|s^t, h^t)$ again is essentially the observation probability $\mathcal{P}^O(\boldsymbol{o}^{t^{-}}|s^t)$. This is not conditioned on the immediate joint actions, because the joint actions have not been taken yet, which is simply a nuance in differential games. The first term $p(\boldsymbol{a}^t|\boldsymbol{o}^{t^{-}}, s^t, h^t)$ literally means the probability of joint actions given joint observation, and the world states, which is essentially related to the joint policies. In order to further factorize this term so as to relate it to the joint policies, we make the assumption that each agent $i$ makes its own decision based on its own type variable $h_i^t$, and its immediate observation $o_i^t$, i.e., a non-recurrent policy that directly maps immediate observation to action. This is a reasonable assumption in a lot of scenarios where agents have good observability such that they do not need to infer a lot of hidden information. For example, in our AIIG, since the opponents have full observability over the type of the protagonist agent, the opponent does not need to hold a belief if the physical states are also observable to it. Based on this mild assumption, we have the following factorization,

\begin{equation}
	p(\boldsymbol{a}^t|\boldsymbol{o}^{t^{-}}, s^t, h^t) = p(\boldsymbol{a}^t|\boldsymbol{o}^{t^{-}}, h^t) \approx \prod_{j}^N \pi_j(o^t_j|h_j)
\label{approximation}
\end{equation} 

To summarize, Eq. (\ref{first_term}) can be represented as:

\begin{equation}
	p(o^t_{i}| s^t, h^t, o^{0:t-1}_i) = \mathbb{E}_{\boldsymbol{a}^t \sim \boldsymbol{\pi}(\bar{\boldsymbol{o}}|\boldsymbol{h})} [\mathcal{P}^O(o^t_{i}| \boldsymbol{a}^t, s^t)],
	\label{simplified_expectation}
\end{equation}

where $\bar{\boldsymbol{o}} = \int \mathcal{P}^O(\boldsymbol{o}^{t^{-}}|s^t) d\boldsymbol{o}^{t^-}$.

The interpretation of Eq. (\ref{simplified_expectation}) is very intuitive: the probability of receiving an observation $o^t_i$ is the expected observation by marginalizing out all the probable joint actions over the observation probability $\mathcal{P}^O(o^t_{i}| \boldsymbol{a}^t, s^t)$, where the probability of the joint actions $\boldsymbol{\pi}(\bar{\boldsymbol{o}}|\boldsymbol{h})$is obtained from the joint policies, by first predicting the expected joint observation of all the agents, and passing it to the joint policies.

The second term in Eq. (\ref{belief_update}), $p(s^t, h^t|o_i^{0:t-1})$ can be expressed as

\begin{equation}
	\begin{aligned}
	p(s^t, h^t|o_i^{0:t-1}) & = \int p(s^t, h^t|s^{t-1}, h^{t-1}) p(s^{t-1}, h^{t-1}|o_i^{0:t-1}) ds^{t-1} dh^{t-1} \\
	& = \int p(s^t, h^t|s^{t-1}, h^{t-1}) b_i^{t-1} ds^{t-1} dh^{t-1}.
	\end{aligned}
	\label{transition_dynamics}
\end{equation}

In order to further simplify this expression, we make a reasonable assumption that $p(s^t, h^t|s^{t-1}, h^{t-1}) = p(s^t, h^t|s^{t-1}, h^{t-1}) \delta(h^t|h^{t-1})$, where $\delta(h^t|h^{t-1})$ denotes the Dirac-delta measure. This assumption means the type of the agents does not change with time, which is a fair assumption in real world applications.  Based on this assumption, Eq. (\ref{transition_dynamics}) simplifies to 

\begin{equation}
	\begin{aligned}
	p(s^t, h^t|o_i^{0:t-1}) = \int p(s^t|s^{t-1}, h^{t}) b_i^{t-1} ds^{t-1}.
	\end{aligned}
	\label{transition_dynamics_simplified}
\end{equation}

The interpretation of Eq. \ref{transition_dynamics_simplified} is pretty clear: the probability over the joint state conditioned on all the previous observation can be predicted based on the posterior belief from last step $b_i^{t-1}$, and keep the belief over agent type unchanged (because the current observation has not been received yet) and propagating the physical states according to some environment dynamic model.  

Combining Eq. \ref{belief_update}, Eq. \ref{simplified_expectation}, and Eq. \ref{transition_dynamics_simplified}, we have the following recursive belief update rule,

\begin{equation}
	b^t_i \propto \mathbb{E}_{\boldsymbol{a}^t \sim \boldsymbol{\pi}(\bar{\boldsymbol{o}}|\boldsymbol{h})} [\mathcal{P}^O(o^t_{i}| \boldsymbol{a}^t, s^t)] \int p(s^t|s^{t-1}, h^{t}) b_i^{t-1} ds^{t-1},
	\label{summary_belief_update}
\end{equation}
where the interpretation is: to infer the state of current step, we can predict it based on the posterior belief of the last step, by propagating the physical state distribution and correcting the belief over the hidden type variable via matching the actual observation with the anticipated observation according to agent policy modeling.

\begin{remark}
In Eq. \ref{summary_belief_update}, the inference over the hidden type variable is implicit inside the expectation term. The observation probability is crucial to the discriminative power of this inference. To illustrate this point, let us consider one extreme case where the observation contains no information about agents' actions, i.e., $\mathcal{P}^O(o^t_{i}| \boldsymbol{a}^t, s^t)$ is not a function of $\boldsymbol{a}^t$. In this case, this expectation term will be independent of the joint policy (will be a constant due to normalization condition of expectation). As a result, no information about the hidden type variable can be extracted from this term. This makes sense, because if the observation tells us nothing about the actions taken by the other agents (dictated by their policies and hidden types), then it is impossible to update our belief over their hidden types. In contrast, suppose the observation contains full information about the joint action (e.g., the protagonist directly observes the joint action), this expectation term would be highly dependent on the joint policies (therefore, on the hidden type variable), and the discriminative power of this inference scheme is maximized.
\end{remark}
  
In order to implement the belief update Eq. \ref{summary_belief_update}, it requires us to know the observation probability $\mathcal{P}^O(o^t_{i}| \boldsymbol{a}^t, s^t)$ and model the state transition probability $p(s^t|s^{t-1}, h^{t})$, and agent policies $\boldsymbol{\pi}$, which is anticipated. Nonetheless, the integral involved in Eq. \ref{summary_belief_update} intractable. 

In this work, we focus on a special case where the physical states are fully observable to all the agent, such that agents do not need to maintain a belief over $s^t$, which significantly simplifies the computational aspect of the problem. Nonetheless, this does not diminish the difficult of the problem, which lies in inferring the hidden type of opponent.

Next, we discuss how to approximate the policies of agent $j$ of each possible type $\{h_j^{(m)}\}_{m=1}^M$. Recall that in the ensemble training step, we create $K$ different policies $\{\pi^{(k)}_{j,m}\}^K_{k=1}$ for each agent of each type. Here we use shorthand $\pi_{j,m}$ to denote agent $j$ with type $h_j^{(m)}$. Each policy within one ensemble can be interpreted as one of the likely strategies that could be adopted by agent $j$ with type $h_j^{(m)}$. However, in the belief update equation, we need only one single policy for agent $j$ with type $h_j^{(m)}$. As a result, we need to synthesize the policy ensemble into one representative policy that can best represent the average behavior of the policy ensemble. We propose to learn this representative policy by minimizing information theoretic distance between this policy and the policy ensemble. More specifically, we choose the Kullback$-$Leibler (KL) divergence as the distance measure, and formulate the following minimization objective function for learning the representative policy $\pi^0_{j,m}$,
\begin{equation}
\begin{aligned}
    J(\pi^0_{j,m}) & = \sum_{k = 1}^K \mathbb{KL}(\pi^{(k)}_{j,m}, \pi^0_{j,m})\\
    & = \sum_{k = 1}^K \mathbb{E}_{\pi^{(k)}_{j,m}}\left[\log \pi^{(k)}_{j,m}-\log \pi^0_{j,m}\right].
\end{aligned}
\label{distillation}
\end{equation}
Eq. \ref{distillation} is essentially policy distillation \cite{distral}. The solution to this minimization is,
\begin{equation}
    \pi^0_{j,m} = \frac{1}{K} \sum_{k = 1}^K \pi^{(k)}_{j,m},
\label{solution2distral}
\end{equation}
which happens to be a simple average over the policies within one ensemble. Conceptually, this is straightforward to implement. However, computationally, averaging $K$ policies is undesirable, because $K$ could be large. Instead, we propose to store an additional action probability term into the shared experience replay buffer, and fit a policy network $\hat{\pi}^0_{j,m; \theta}$ to samples of action probability from the experience replay using mean square error (MSE) loss. This operation approximates Eq. \ref{solution2distral}, but at almost constant computational complexity, since we do not need any additional computations to obtain the action probability sample.

\subsection{Policy ensemble optimization}

The ensemble training step typically improves the robustness of the protagonist agent's policy. However, two problems need to be addressed to make this approach more effective and efficient. First, we want a metrics for measuring policy robustness and we want to explicitly optimize this robustness metrics. Second, we want to minimize the additional computation overhead introduced by ensemble training. 

We propose to address these two problems through a meta-optimization of the policy ensemble. Instead of using a fixed-size ensemble, we dynamically resize the ensemble through three operations: \textbf{pop}, \textbf{append} and \textbf{exchange}. \textbf{pop} randomly removes one policy from the ensemble and push it into a deactivation-set. \textbf{append} randomly selects one policy from the deactivation-set and append it to the ensemble. \textbf{exchange} randomly selects one policy from both the ensemble and the deactivation-set and exchanges them with each other. 

The objective of modifying the ensemble is to obtain a good trade-off between robustness and computational complexity, which is dominated by the ensemble size. We propose to measure the robustness via Procedure 1,
\begin{algorithm}[t]
\floatname{algorithm}{Procedure}
\caption{Ensemble evaluation}
\begin{algorithmic}[1]
%\Procedure{Euclid}{$a,b$}\Comment{The g.c.d. of a and b}
\State Fix the protagonist policy
%\While{$r\not=0$}\Comment{We have the answer if r is 0}
\State Train a single opponent policy against the fixed protagonist policy
\State Obtain the average protagonist agent reward $r^p$ and opponent reward $r^o$ after training
%\EndWhile\label{euclidendwhile}
%\State \textbf{return} $b$\Comment{The gcd is b}
%\EndProcedure
\end{algorithmic}
\label{procedure_evaluation}
\end{algorithm}

and we define the following metrics,
\begin{equation}
    \rho = -r^p + \lambda_1 r^o + \lambda_2 K,
\end{equation}
where $\lambda_1, \lambda_2$ are weight parameters, and $K$ is the varying size of the policy ensemble.

The combined reward term $-r^p + \lambda_1 r^o$ is a measure of the robustness of the protagonist policy, which is noisy due to the intrinsic stochasticity of reinforcement learning, while $K$ is a surrogate measure for computation complexity. Therefore, minimizing $\rho$ leads to an optimal trade-off between policy robustness and computation complexity. We interpret this minimization problem as a stochastic optimization over the powerset of the initial policy ensemble. We solve this stochastic optimization via simulated annealing as described in Procedure~2.

\begin{algorithm}[t]
\floatname{algorithm}{Procedure}
\caption{Ensemble Optimization}
\begin{algorithmic}[1]
\State Randomly select an operation $\xi$ from \{\textbf{pop}, \textbf{append}, \textbf{exchange}\} to apply on the policy ensemble
\State Obtain a new metrics $\rho_{\text{new}}$ via Procedure \ref{procedure_evaluation}
\State Accept the operation $\xi$ with probability $p$, where $p = \exp (\min\{0, \rho_{\textbf{old}} - \rho_{\textbf{new}}\} / T)$
\end{algorithmic}
\label{procedure_evaluation}
\end{algorithm}

\section{Evaluation}
This section addresses the following questions:
\begin{enumerate} \itemsep -0.025in
    \item Is it necessary to use ensemble training, considering its additional computation overhead?
    \item Is it beneficial to explicitly model opponent policy and maintain a belief?
    \item How much improvement do we get from ensemble training and ensemble meta-optimization?
\end{enumerate}

\subsection{Scenario: two-player asymmetric game}

\begin{figure}[t]
  \centering
  \includegraphics[trim=10 0 0 10,clip,width=0.4\columnwidth]{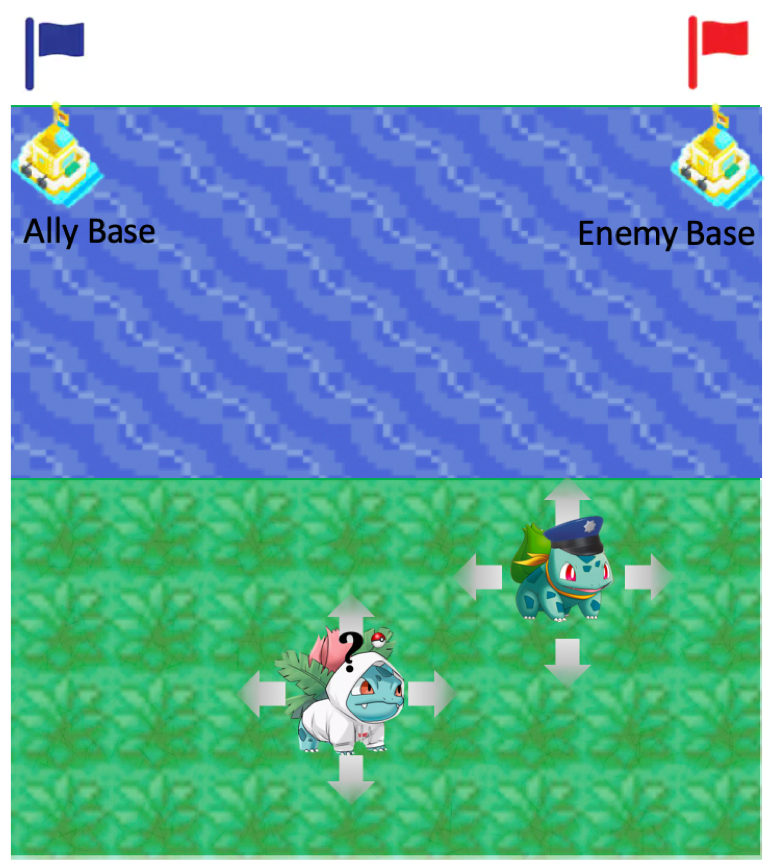}
  \caption{Asymmetric two player imperfect-information scenario in a 2-d world: the protagonist is a grasshopper officer, while the opponent could be a beaver (ally) or a turtle (enemy). The protagonist does not know the actual type of the opponent because of the white cloak.}
  %\vspace{-0.1in}
\label{scenario} 
\end{figure}

We design a two-player asymmetric-information game to evaluate our algorithm, as illustrated in Fig. \ref{scenario}. There are two agents: the protagonist agent (a grasshopper officer) and the opponent agent with two possible types (either an ally beaver or an enemy turtle). The opponent's objective is to reach its home base (depending on its type) as soon as possible. The protagonist's objective is to identify the type of its opponent, and obtain reward by tagging the opponent if it turns out to be an enemy turtle. Mistakenly tagging an ally beaver would incur a large penalty to the protagonist. The grasshopper protagonist does not swim, so once the opponent jumps into the river, the officer cannot tag it anymore. If the opponent is an enemy turtle, it receives large penalty if tagged. The opponent always receives penalty if it has not reached its base, and the penalty increases with its distance from its base. 

The detailed description of this game and the hyper-parameters are provided in the supplement materials.

\subsubsection{Description of evaluation domain}

The game domain is a $8\times 8$ continuous square area. At the beginning of each game episode, the opponent starts from the bottom middle of the world. The opponent type is randomly sampled with equal probability from the two possible types (ally beaver and enemy turtle).  

\subsubsection{State and action space}

The state of each agent is its 2-d position, i.e., $\mathcal{S}_i = [0, 8] \times [0, 8]$. The protagonist agent has a discrete action space $\mathcal{A}^p = [\textbf{move left}, \textbf{move right}, \textbf{move up},$ $\textbf{move down}, \textbf{tag}, \textbf{probe}]$, and the opponent agent's action space is $\mathcal{A}^o = [\textbf{move left}, \textbf{move right}, \textbf{move up}, \textbf{move down}]$. Each of the `move' action changes the agent position by one unit distance. The tag action succeeds if and only if the distance between the two agents is less than 2.5. The probe action is equivalent to query a noisy measurement of the opponent's true type, where there is 0.8 probability getting the correct type and 0.2 probability getting the wrong type. The protagonist agent could take this probe action to help with its inference besides simply observing the opponent. Each probe action incurs cost, so the protagonist agent has to wisely decide when and how many times to probe.

\subsubsection{Reward}

The reward of the opponent agent consists of two parts: (1) $r_d = -0.25d^{2/5}$, where $d$ is its distance from its home base; (2) $r_{\text{tagged}} = -10$ if being tagged. 

The state-action reward of the protagonist agent consists of several parts: (1) $r_{\text{tag enemy}} = 10$ if tagging an enemy; (2) $r_{\text{tag ally}} = -20$ if tagging an ally; (3) $r_{\text{d2o}} = -0.25d_{\text{o}} ^{2/5}$, where $d_{\text{o}}$ is the distance between the protagonist agent and the opponent. This is a heuristic reward to help the protagonist agent learning sensible behaviors; (4) Tag cost $r_{\text{tag cost}} = -0.2$, no matter tagging is successful or not; (5) Probe cost $r_{\text{probe cost}} = -0.25C$, where $C$ is the cumulative counts of the probing action so far, i.e., the probe cost per time increases as the total number of probing increases. This effectively prevents the agent from abusing the probe action.

Based on the rule of this game, an enemy turtle might take multiple different strategies. For example, one strategy is to rush towards its home base to minimize the distance penalty. However, the protagonist can quickly identify the enemy and try to tag it. As a result, the enemy might end up getting a huge penalty as being tagged. Another strategy is to initially head towards the ally base, such that the officer would be fooled to believe that the opponent is an ally. Once the enemy is close enough to the river, it can jump into the river and rush to its base. This strategy incurs larger distance penalty, but eventually might get a higher reward by avoiding being tagged. 

\subsubsection{Learning algorithm and ensemble optimization hyper-parameters}

The reinforcement learning algorithm we used to train the opponent is TD3. The hyper-parameter of the algorithim is listed in Table 

\begin{table}[ht] 
\caption{Hyper-parameter of TD3} % title of Table 
\centering      % used for centering table 
\begin{tabular}{c c}  % centered columns (4 columns) 
\hline\hline                        %inserts double horizontal lines 
Hyper-parameter & Value\\ [0.3ex]
\hline
Actor learning rate & 5e-5\\ [0.3ex]
Critic learning rate & e-3\\ [0.3ex]
Target net soft-update rate & 5e-3\\ [0.3ex]
Exploration noise std & 0.2\\ [0.3ex]
Noise clip threshold & 0.5\\ [0.3ex]
\hline     %inserts single line 
\end{tabular}
\label{hyper-parameter}
\end{table}

\vspace{-0.2in}

\begin{table}[ht] 
\caption{Hyper-parameter of ensemble optimization} % title of Table 
\centering      % used for centering table 
\begin{tabular}{c c}  % centered columns (4 columns) 
\hline\hline                        %inserts double horizontal lines 
Hyper-parameter & Value\\ [0.3ex]
\hline
Opponent loss weight $\lambda_1$ & 0.1\\ [0.3ex]
Ensemble size weight $\lambda_2$ & 1.0\\ [0.3ex]
Initial temperature $T_0$ & 30.0\\ [0.3ex]
Minimum temperature $T_{\text{min}}$ & 0.2\\ [0.3ex]
Temperature decay rate & 0.975\\ [0.3ex]
\hline     %inserts single line 
\end{tabular}
\label{hyper-parameter}
\end{table}

\subsection{Ensemble training vs. single model}
To answer the first question, we compared the protagonist policy learned from training against an ensemble of opponent policies and that from training against a single opponent policy. We used a similar ensemble as used in \cite{cerl}, which consists of four opponent policies, each policy is learned from training against the protagonist policy. We used four different discount factors for the opponent learning objectives: $\gamma_1 = 0.9, \gamma_2 = 0.99, \gamma_3 = 0.997, \gamma_4 = 0.9995$. An interpretation of this setting is a variety of opponent playing styles ranging from myopic to far-sighted strategies.  

For comparison, we also trained the protagonist policy individually against each opponent model, so we obtained five protagonist policies in total. For evaluation, we trained five separate opponent evaluation policies, each corresponding to one of the protagonist policies. The evaluation policies all used the same discount factor $\gamma = 0.99$.

Fig. \ref{blue_red_rewards_evol} shows the training and evaluation rewards. During training, the single model policies generally lead to higher protagonist reward, while the ensemble training results in the lowest protagonist reward. This suggests that the protagonist policy overfits to one of the single opponent models, thus achieving high training reward but low evaluation reward. In contrast, the protagonist policy trained against the ensemble achieves the best evaluation reward. It is worth pointing out that, in the second single model setting, although the hyper-parameter $\gamma_2 = 0.99$ is exactly the same as that of the evaluation opponent, the evaluation reward is still significantly worse than the training reward. This is not surprising, as agent could learn different policies even with the same hyper-parameter setting. Therefore, overfitting is almost inevitable when training against single model.

\subsection{Belief space policy vs.~RNN}

\begin{figure*}[p]
	\centering 
	\subfigure[Protagonist training reward: higher is better] { 
		\label{evo_blue} 
		\includegraphics[width=0.48\columnwidth]{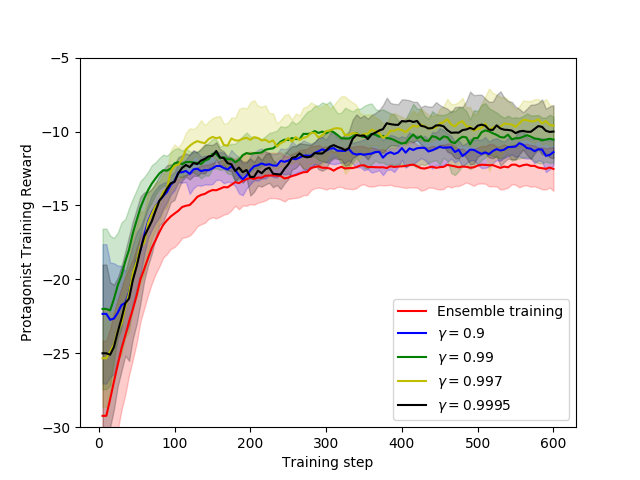} 
	} 
	\subfigure[Protagonist evaluation reward: higher is better] { 
		\label{evo_red} 
		\includegraphics[width=0.48\columnwidth]{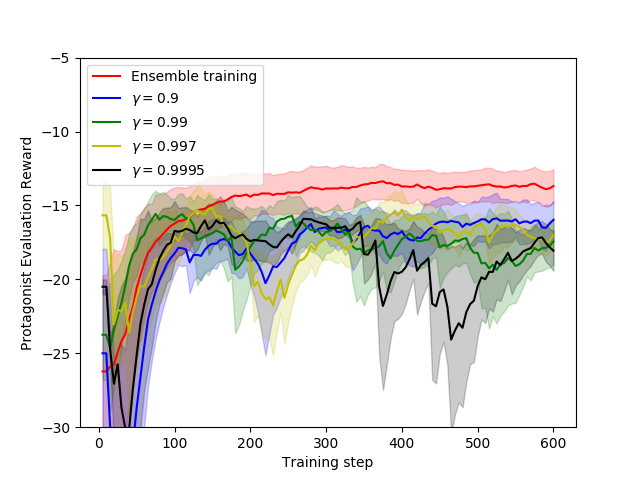} 
	} 
	\subfigure[Opponent (enemy type) training reward: lower is better] { 
		\label{evo_blue} 
		\includegraphics[width=0.48\columnwidth]{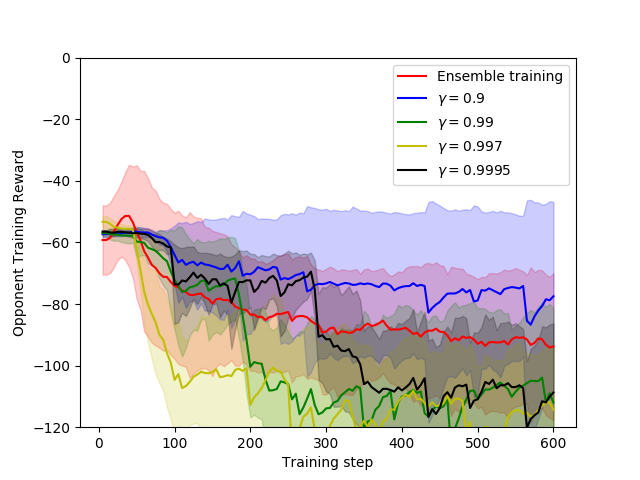} 
	} 
	\subfigure[Opponent (enemy type) evaluation reward: lower is better] { 
		\label{evo_red} 
		\includegraphics[width=0.48\columnwidth]{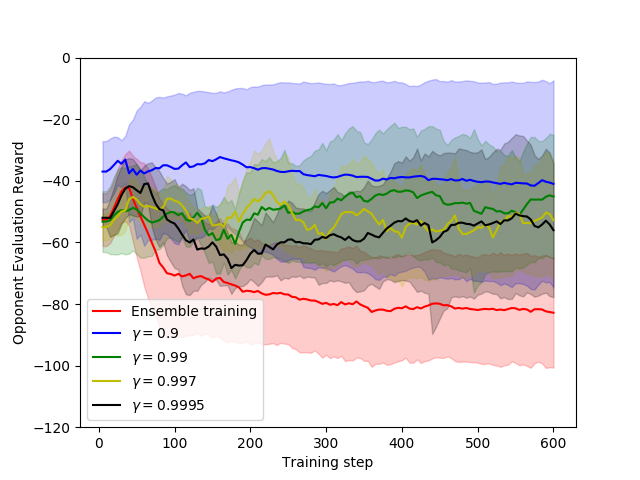} 
	} 
	%\vspace{-0.1in}
	\caption{Training and Evaluation rewards of the protagonist agent and the opponent agent: Single opponent models ({\color{blue}$\gamma = 0.9$}, {\color{ForestGreen}$\gamma = 0.99$}, {\color{Dandelion}$\gamma = 0.997$}, {\color{black}$\gamma = 0.9995$}) performs better than {\color{red}ensemble training} in the training phase due to overfitting to simple opponent models, while ensemble training outperforms single opponent models in evaluation} 
	\label{blue_red_rewards_evol} 
%\end{figure*}
%\begin{figure*}[p]
	\centering 
	\subfigure[Protagonist reward: higher is better] { 
		\label{rnn_blue} 
		\includegraphics[trim=10 0 0 30,clip,width=1\columnwidth]{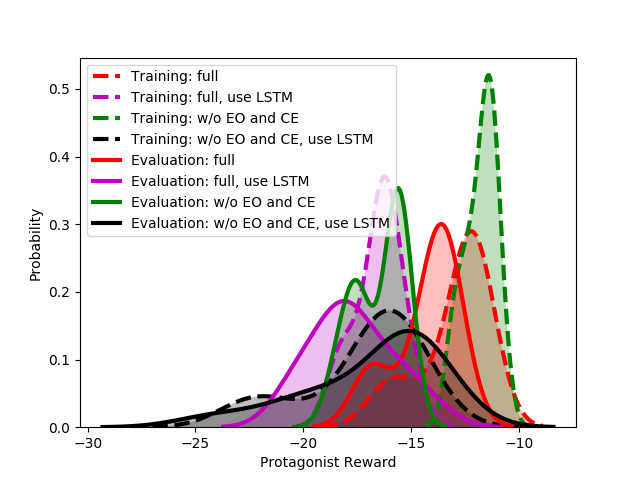} 
	} 
	\subfigure[Opponent (enemy type) reward: lower is better] { 
		\label{rnn_red} 
		\includegraphics[trim=10 0 0 30,clip,width=1\columnwidth]{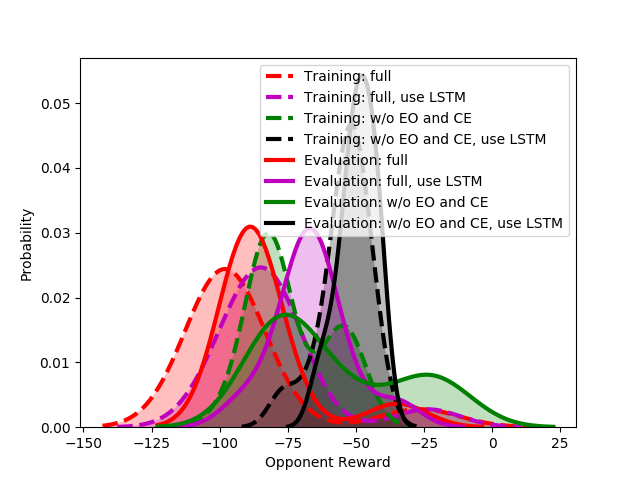} 
	} 
	%\vspace{-0.1in}
	\caption{Training and Evaluation rewards of the protagonist agent (left) and the opponent agent (right): {\color{red}\textbf{(1)}} belief space policy, with ensemble optimization (EO) and cooperative-evolution (CE); {\color{Fuchsia}\textbf{(2)}} LSTM, with EO and CE; {\color{ForestGreen}\textbf{(3)}} belief space policy, without EO and CE (single opponent model); {\color{black}\textbf{(4)}} LSTM, without EO and CE (single opponent model); Belief space policy \{{\color{red}\textbf{(1)}} and {\color{ForestGreen}\textbf{(3)}}\} outperforms LSTM \{{\color{Fuchsia}\textbf{(2)}} and {\color{black}\textbf{(4)}}\}} 
	\label{cmp_with_rnn} 
%\end{figure*}
%\begin{figure*}[h!] 
\centering 
    \subfigure[Protagonist reward: higher is better] { 
    \label{abl_blue} 
    \includegraphics[trim=10 0 0 30,clip,width=1\columnwidth]{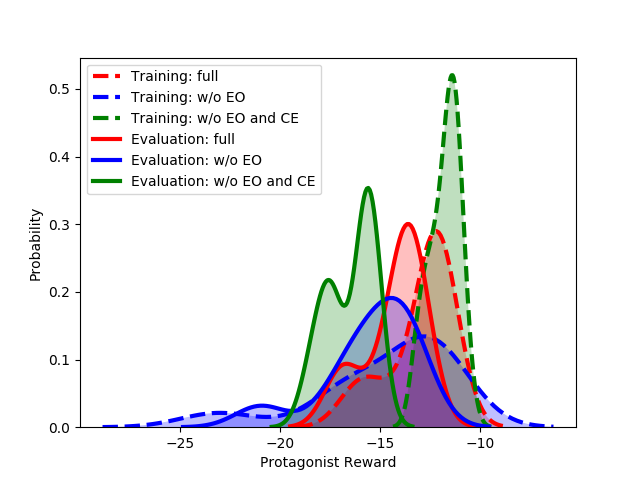} 
    } 
    \subfigure[Opponent (enemy type) reward: lower is better] { 
    \label{abl_red} 
    \includegraphics[trim=10 0 0 30,clip,width=1\columnwidth]{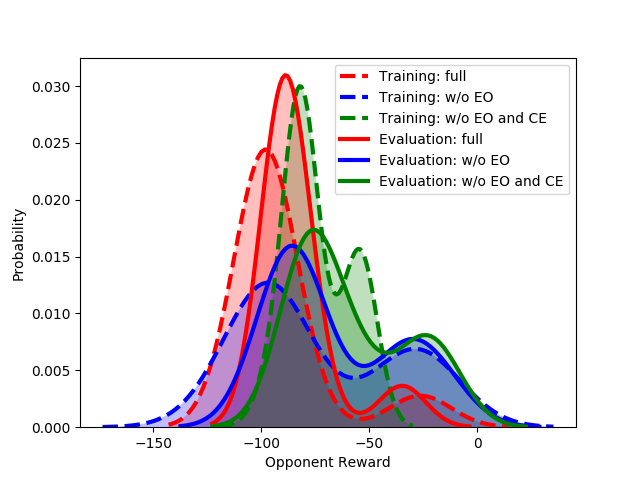} 
    } 
   % \vspace{-0.1in}
\caption{Training and Evaluation rewards of the protagonist agent (left) and the opponent agent (right): {\color{red}\textbf{(1)}} with both ensemble optimization (EO) and cooperative-evolution (CE); {\color{blue}\textbf{(2)}} with CE but without EO; {\color{ForestGreen}\textbf{(3)}} without EO and CE (single opponent model); \textbf{{\color{red}EO + CE}} outperforms \textbf{{\color{blue}CE only}}, which outperforms \textbf{{\color{ForestGreen} single opponent model}}} 
\label{ablation_study} 
\end{figure*}

To answer the second question, we replaced belief space policy with a recurrent policy parameterized by a LSTM. Fig. \ref{cmp_with_rnn} and Table \ref{vs_lstm_mean}
show the comparison between these two settings, where the belief space policy consistently outperforms the recurrent policy. This result agrees with our conjecture that learning recurrent policy might be difficult due to lack of prior knowledge on the information structure and the high-variance state-space reward.
%\vspace{-0.1in}
\begin{table}[h] 
\caption{Mean reward: v.s. LSTM (Training / Evaluation)} % title of Table 
\centering      % used for centering table 
\begin{tabular}{l c c c}  % centered columns (4 columns) 
\hline\hline                        %inserts double horizontal lines 
Algorithm &Protagonist & Enemy\\ [0.3ex]
\hline
\text{{\color{red}belief space, with EO \& CE}} & -13.2 / \textbf{-14.4} & \textbf{-90.8} / \textbf{-83.0}\\
\text{{\color{Fuchsia} LSTM, with EO \& CE}} & -16.5 / -17.7 & -80.6 / -66.2\\ [0.3ex]
\text{{\color{ForestGreen}belief space, w/o EO \& CE}} & \textbf{-11.8} / -16.5 & -73.8 / -58.6\\ [0.3ex]
\text{{\color{black} LSTM, w/o EO \& CE}} & -17.2 / -16.8 & -54.2 / -49.4\\ [0.3ex]
\hline     %inserts single line 
\end{tabular}
\label{vs_lstm_mean}
\end{table}
%\vspace{-0.1in}

\subsection{Ablation study}

To answer the third question, we compared our algorithm with its ablated versions: (I) without neuro-evolution, (II) without both neuro-evolution and ensemble optimization. For the ablated version II, we randomly sampled subsets of the ensemble from its powerset, and used the fixed subset for training. Fig. \ref{ablation_study} and Table \ref{abl_mean} show the training and evaluation rewards of the full and ablated versions of our algorithm. The result suggests that both neuro-evolution and ensemble optimization have important contribution to the performance improvement.
%\vspace{-0.1in} 
\begin{table}[h!] 
\caption{Mean reward: ablation study (Training / Evaluation)} % title of Table 
\centering      % used for centering table 
\begin{tabular}{c c c c}  % centered columns (4 columns) 
\hline\hline                        %inserts double horizontal lines 
Ablated version &Protagonist & Enemy\\ [0.3ex]
\hline
\text{{\color{red}with EO \& CE}} & -13.2 / \textbf{-14.4} & \textbf{-90.8} / \textbf{-83.0}\\
\text{{\color{blue}w.o EO}} & -15.0 / -15.6 & -73.6 / -65.8\\ [0.3ex]
\text{{\color{ForestGreen}w.o EO \& CE}} & \textbf{-11.8} / -16.5 & -73.8 / -58.6\\ [0.3ex]
\hline     %inserts single line 
\end{tabular}
\label{abl_mean}
\end{table} 
%\vspace{-0.1in}
\section{Summary}
We summarize the key findings of this work as follow:
\begin{itemize}\itemsep -.025in
    %\item We propose a new decision-making framework, POSGEOS modified from POSG, which is suitable for reasoning in scenarios involving uncertain opponent types. %\XX{ok, but say why this is an important thing to have - is it important for us or useful to others?}
    \item We propose algorithms based on MARL and ensemble training for robust opponent modeling and posterior inference over the opponent type from the observed action.
    \item We propose an explicit metrics for policy robustness evaluation, and formulate a stochastic optimization to maximize robustness and minimize computation complexity.
    \item We empirically demonstrate that the  explicit opponent modeling outperforms a black-box RNN approach, and the stochastic optimization results in better results (in terms of the robustness-complexity trade-off) than standard ensemble training approach. %\XX{is this what you did? not obvious from the wording}
    
\end{itemize}

%\begin{acks}

%The authors want to thank Dongki Kim, Kasra Khosoussi, and Chuangchuang Sun for their help and insightful discussions. This work is supported by ARL DCIST under Cooperative Agreement Number W911NF-17-2-0181, and research agreement with SSCI agreement number SC-1661-04, which is under prime contract with DARPA Contract 140D6319C0005.

%\end{acks}

\section*{Acknowledgement}
This work is supported by ARL DCIST
under Cooperative Agreement Number W911NF-17-2-0181, Scientific Systems Company, Inc. under research agreement $\#$ SC-1661-04 and computation support through Amazon Web Services. The authors would like to thank Kasra Khosoussi, Dongki Kim, and Chuangchuang Sun for the insightful discussions.

%%%%%%%%%%%%%%%%%%%%%%%%%%%%%%%%%%%%%%%%%%%%%%%%%%%%%%%%%%%%%%%%%%%%%%%%%%%%%%%%%%%%%%%%%%%%%%%%%%%%%%%%%
%% bibliography: see CFP for number of permitted pages

\bibliographystyle{ACM-Reference-Format}  % do not change this line!
\bibliography{ref}  % put name of your .bib file here

\end{document}